\documentclass[letterpaper, 10 pt, conference]{ieeeconf}  

\IEEEoverridecommandlockouts                              

\usepackage{graphicx}
\usepackage{amssymb}
\usepackage{bm}
\usepackage{cite}
\usepackage{amsmath}
\usepackage{subfigure}
\usepackage{stfloats}
\usepackage{multirow}
\usepackage{gensymb}
\usepackage{xcolor}
\usepackage{boldline}
\usepackage{cuted}
\usepackage[T1]{fontenc}
\usepackage[final]{changes}
\usepackage{threeparttable}
\usepackage[colorlinks=true, allcolors=black]{hyperref}

\title{\LARGE \bf
	Fast Extrinsic Calibration for Multiple Inertial Measurement Units in Visual-Inertial System
}

\author{Youwei Yu$^{1}$, Yanqing Liu$^{1,2}$, Fengjie Fu$^{1,2}$, Sihan He$^{1}$, Dongchen Zhu$^{1,2}$,\\Lei Wang$^{1,2}$, Xiaolin Zhang$^{1,2,3,4,5}$, Jiamao Li$^{1,2,4}$
	\thanks{$^{1}$ Bionic Vision System Laboratory, State Key Laboratory of Transducer Technology, Shanghai Institute of Microsystem and Information Technology, Chinese Academy of Sciences, Shanghai 200050, China.}%
	\thanks{$^{2}$ University of Chinese Academy of Sciences, Beijing, 100049, China.}%
	\thanks{$^{3}$ ShanghaiTech University, Shanghai, 201210, China.}%
	\thanks{$^{4}$ Xiongan Institute of Innovation, Xiongan, 071700, China.}%
	\thanks{$^{5}$ University of Science and Technology of China, Hefei, 230027, China.}%
	\thanks{Corresponding author: Jiamao Li, \href{mailto:jmli@mail.sim.ac.cn}{\texttt{jmli@mail.sim.ac.cn}} }%
	\thanks{This work was supported in part by the National Science and Technology Major Project from Minister of Science and Technology of China(2018AAA0103100), in part by National Natural Science Foundation of China (62003326), in part by the Shanghai Municipal Science and Technology Major Project (ZHANGJIANG LAB) under Grant 2018SHZDZX01, in part by Youth Innovation Promotion Association, in part by Chinese Academy of Sciences(2021233), and in part by Shanghai Academic Research Leader(22XD1424500).}
}

\begin{document}

	\maketitle

	\begin{abstract}
		
		In this paper, we propose a fast extrinsic calibration method for fusing multiple inertial measurement units (MIMU) to improve visual-inertial odometry (VIO) localization accuracy. Currently, data fusion algorithms for MIMU highly depend on the number of inertial sensors. Based on the assumption that extrinsic parameters between inertial sensors are perfectly calibrated, the fusion algorithm provides better localization accuracy with more IMUs, while neglecting the effect of extrinsic calibration error. Our method builds two non-linear least-squares problems to estimate the MIMU relative position and orientation separately, independent of external sensors and inertial noises online estimation. Then we give the general form of the virtual IMU (VIMU) method and propose its propagation on manifold. We perform our method on datasets, our self-made sensor board, and board with different IMUs, validating the superiority of our method over competing methods concerning speed, accuracy, and robustness. In the simulation experiment, we show that only fusing two IMUs with our calibration method to predict motion can rival nine IMUs. Real-world experiments demonstrate better localization accuracy of the VIO integrated with our calibration method and VIMU propagation on manifold. \\
		
	\end{abstract}
	
	

	\section{INTRODUCTION}
	
	\PARstart{I}{n} the realm of simultaneous localization and mapping (SLAM), multiple inertial sensors fusion is a popular topic. IMU, as a proprioceptive sensor, outputs high-frequency measurements but suffers from pose drift in long-term integration \cite{c1,c2,c3,c4,c5,c6,c8}. Thus, IMU can cooperate with exteroceptive sensors such as the LiDAR and camera that provide global observation. Given the compact size and low cost of the microelectromechanical  systems (MEMS) IMU, the SLAM system can add more inertial sensors for failure detection or measurements fusion.
	
	\begin{figure}[thpb]
		\centering
		\includegraphics[scale=0.24]{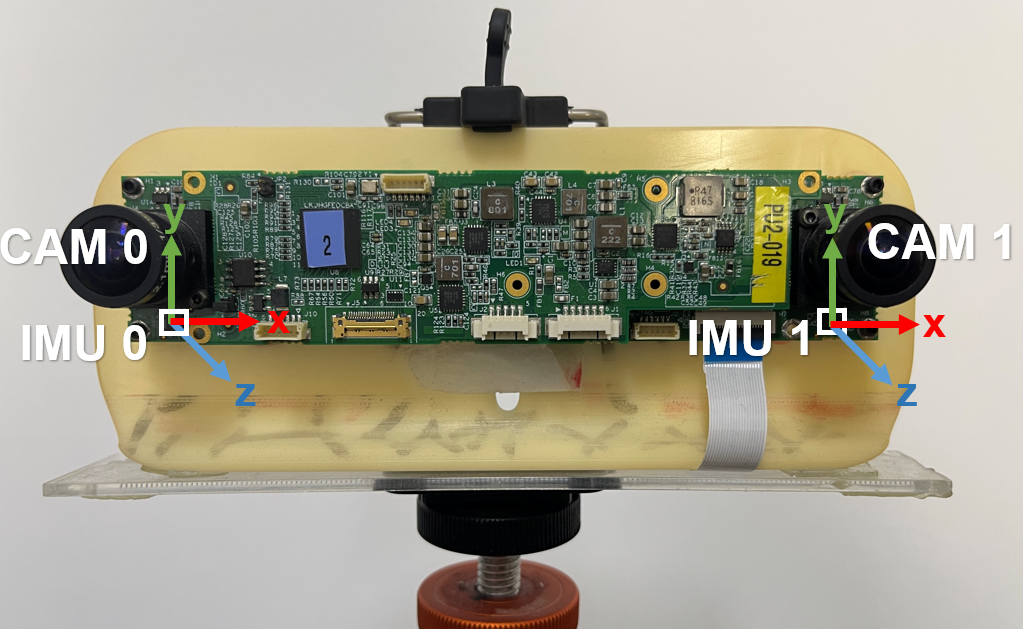}
		\caption{The binocular and two-IMU sensor board used for extrinsic calibration comparison and VIO experiments. Stereo images are synchronized, and inertial measurements have exact timestamps.}
		\label{leadsense}
	\end{figure}
	
	Extrinsic accuracy of MIMU is seldom discussed, while it plays a significant role in fusion algorithms. Most visual-MIMU systems assume the relative pose between each inertial sensor and the body is perfectly calibrated. However, our simulation experiments show that only one IMU can provide better preintegration accuracy if the system cannot guarantee precise extrinsic parameters.
	
	So far, existing extrinsic calibration methods for MIMU need the exact trajectory of the body, either estimated by the expensive turntable \cite{c18} or exteroceptive sensors (e.g., Kalibr \cite{c15, c16}). Although these algorithms can perform very well in specified environments, the calibration method independent of environment restrictions or additional devices is more practicable. Meanwhile, the method free from sensor noises online estimation remains a challenge, limiting calibration accuracy and computation efficiency.
	
	Our proposed method builds two non-linear least-squares problems to estimate relative translation and orientation parameters between inertial sensors. First, we optimize the relative orientation based on raw gyroscope measurements. Second, inspired by the virtual IMU method \cite{c6}, we generate angular acceleration measurements with less noise to enhance the relative position calibration performance. Note that our method does not rely on the ground-true trajectory or external sensors. Meanwhile, we treat the online estimation of inertial noise terms as an impairment of accuracy to avoid the problem of over-fitting.
	
	Specifically, the main contributions of this paper include:
	\begin{itemize}
		\item We propose a fast extrinsic calibration method between MIMU. We validate on datasets \cite{c19}, our self-made sensor board, and the sensor suit with RealSense T265 and D435i, showing that the proposed method outperforms Kalibr and the method in \cite{c19} regarding the accuracy, computation efficiency, and robustness.
		
		\item Simulation experiment demonstrates that only integrating two IMUs with our method to predict motion can rival nine IMUs.
		
		\item We give the general form of the VIMU method and propose its propagation on manifold. Experiments on datasets \cite{c26} and our binocular and two-IMU sensor board (see Fig. \ref{leadsense}) show that our proposed method improves localization accuracy in the VIO system.
	\end{itemize}

	\section{RELATED WORK}
	We group the MIMU research into two aspects: sensor data fusion and extrinsic calibration.
	
	\subsection{MIMU Data Fusion}
	One main focus of MIMU is failure detection and isolation (FDI). Yang \emph{et al.} \cite{c29} proposed the averaged parity vector (AVP) algorithm to enhance isolation identification. Eckenhoff \emph{et al.} \cite{c5} developed a method that could seamlessly choose the base IMU from auxiliary ones. Nonetheless, Bancroft \emph{et al.} \cite{c9} did not recommend FDI, especially for low-cost MEMS IMU in rapid movements, since the measurement model can hardly be zero-mean, violating the FDI premise.
	
	Another main focus is accuracy improvement, including the stacked-state method, VIMU generation, and the best layout. First, Bancroft \emph{et al.} \cite{c9} stacked all inertial measurements into a nine-parameter VIMU least-squares estimator. This method had a large observation size and could not deal with only two IMUs because of rank deficiency.
	
	Second, Zhang \emph{et al.} \cite{c6} proposed a lightweight and accurate VIMU method based on a least-squares estimator. Patel \emph{et al.} \cite{c14} showed that the MIMU GPS-denied with VIMU or Federated Kalman Filter could perform comparably to the single-IMU GPS-aided algorithm. Faizullin \emph{et al.} \cite{c7} dynamically chose three non-coplanar axes among gyroscopes. Although this method \cite{c7} provided excellent orientation estimation, we cannot integrate it into the SLAM system since it did not deal with linear accelerations.
	
	Third, as a comprehensive study on the optimal layout of MIMU \cite{c31,c32}, Zhu \emph{et al.} \cite{c11} proposed the constrained multi-dimensional fruit fly optimization algorithm (CMFOA) to estimate the optimal weight of optimal weighted least-squares (WLS). However, this method is not suitable for our two-IMU sensor board.

	\subsection{MIMU Extrinsic Calibration}
	The MIMU extrinsic calibration falls into two categories. The first was to use the turntable instrument to get the exact trajectory of the IMU array rigid body \cite{c17,c18}. Similarly, some other methods used exteroceptive sensors to estimate trajectory. A state-of-the-art method is Kalibr by Rehder \emph{et al.} \cite{c15,c16}, given cameras and fiducial targets (i.e., checkerboard). On the one hand, this method requires exteroceptive sensors, good illumination, and smooth movement. On the other hand, to get the relative pose between each IMU, we need to multiply transformation matrices, leading to cumulative calibration errors.
	
	Secondly, some methods performed calibration without external sensors. Eckenhoff \emph{et al.} \cite{c5} proposed to minimize preintegration pose error terms based on spatial and temporal relations. Similarly, Kim \emph{et al.} \cite{c20} used the inertial relative preintegration information to build residual functions. Both methods in \cite{c5,c20} stated that they could reduce the calibration error since the noise terms were multiplied by the IMU time step. Nonetheless, the inertial measurements were also multiplied by the time step. Lee et al. \cite{c19} developed a non-linear least-squares optimization problem that surpassed Kalibr in terms of accuracy, success rate, and computation time. However, this method treated IMU noises and angular accelerations as variables to be estimated, ignoring that kinetics restrictions on these terms could be loosened during iterated optimization. As a result, this method may cause the problem of overfitting. In contrast, our proposed method incorporates these factors into the optimization error.
	
	The organization of the rest is as follows. Section \ref{FAST EXTRINSIC CALIBRATION FOR MIMU} presents our fast extrinsic calibration method for MIMU. Section \ref{VIRTUAL IMU METHOD REVISITED} gives the general form and analysis of the VIMU generation method and then derives VIMU propagation equations on manifold. Section \ref{experiments} shows our method's calibration performance with three different sensor boards, pure inertial integration with our method, and localization accuracy when integrated into the VIO system.

	\section{FAST EXTRINSIC CALIBRATION FOR MIMU}
	\label{FAST EXTRINSIC CALIBRATION FOR MIMU}
	
	In this section, we propose a fast extrinsic calibration method for MIMU by estimating relative orientation and translation separately.
	
	\subsection{State Variables and Frames}
	Assume we use two IMUs, $ A $ and $ B $, moving in the world frame $ \{W\} $. IMU frames are denoted by $ \{I\} $ ,$ \{A\} $, $ \{B\} $, and the virtual one $ \{V\} $. Common MEMS IMU outputs 3-axis angular velocities $ {}^{I} \widetilde{\boldsymbol{\omega}} \in \mathbb{R}^3 $ and linear accelerations $ {}^{I} \boldsymbol{a} \in \mathbb{R}^3 $, where the reference frame is $ \{I\} $.
	
	The symbols $ \widetilde{\mathbf{x}} $, $ \widehat{\mathbf{x}} $, $\mathbf{x}$ mean the actual, estimate and ground-true value of sensor measurement $ \mathbf{x} $ respectively. The symbol $ \lfloor \cdot \rfloor $ represents the skew-symmetric matrix of vector $ (\cdot) $.
	
	Rotation matrix $ {}^{A}\mathbf{R}_{V} \in SO(3) $ represents the rotation \deleted{from $ \{A\} $ to $ \{V\} $} \added{from $ \{V\} $ to $ \{A\} $}. Similarly, translation matrix $ {}^{A}\mathbf{p}_{V} \in \mathbb{R}^3 $ represents the linear translation \deleted{from $ \{A\} $ to $ \{V\} $} \added{from $ \{V\} $ to $ \{A\} $}.
	
	\subsection{IMU Measurement Model}
	Give the IMU measurements model:
	\begin{equation}
		\begin{aligned}
			{}^{I} \widetilde{\boldsymbol{\omega}}(t) & = {}^I \boldsymbol{\omega}(t) + \boldsymbol{b}_g(t) + \boldsymbol{\eta}_g(t) \\
			{}^I \widetilde{\boldsymbol{a}}(t)& = {}^{I}\boldsymbol{R}_{W}({}^{W}\boldsymbol{a}(t)- {}^{W}\boldsymbol{g}) + \boldsymbol{b}_a(t) + \boldsymbol{\eta}_a(t)
		\end{aligned}
	\end{equation}
	where
	\begin{itemize}
		\item $ \boldsymbol{b}_g $ and $ \boldsymbol{b}_a $ are biases modeled as random walk
		\begin{equation}
			\begin{aligned}
				& \dot{\boldsymbol{b}}_g(t)  = \boldsymbol{n}_{b_g}(t) \sim \mathcal{N}(\boldsymbol{0}, \boldsymbol{\sigma}_{b_g}^2 \mathbf{I}_3) \\
				& \dot{\boldsymbol{b}}_a(t) = \boldsymbol{n}_{b_a}(t) \sim \mathcal{N}(\boldsymbol{0}, \boldsymbol{\sigma}_{b_a}^2 \mathbf{I}_3) ,
			\end{aligned}
		\end{equation}
		\item $ \boldsymbol{\eta}_g $ and $ \boldsymbol{\eta}_a $ are noises modeled as \deleted{Wiener process}\added{Gaussian noise}
		\begin{equation}
			\begin{aligned}
				& \boldsymbol{\eta}_g \sim \mathcal{N}(\boldsymbol{0}, \boldsymbol{\sigma}_{g}^2 \mathbf{I}_3), \quad \boldsymbol{\eta}_a \sim \mathcal{N}(\boldsymbol{0}, \boldsymbol{\sigma}_{a}^2 \mathbf{I}_3) ,
			\end{aligned}
		\end{equation}
		\item $ {}^{W}\boldsymbol{g} $ is the gravity vector expressed in the world frame.
	\end{itemize}
	
	\subsection{Problem Statement}
	Give the angular speed and specific force relations of two IMUs on a rigid body in the inertial frame:
	\begin{equation}
		\begin{aligned}
			{}^{B} \boldsymbol{\omega} &= {}^{B} \mathbf{R}_{A} {}^{A} \boldsymbol{\omega} \\
			{}^{B} \boldsymbol{a} &= {}^{B} \mathbf{R}_{A} \left( {}^{A} \boldsymbol{a} + \lfloor {}^{A} \boldsymbol{\omega} \rfloor^2 {}^{A} \mathbf{p}_{B} + \lfloor {}^{A} \dot{\boldsymbol{\omega}} \rfloor {}^{A} \mathbf{p}_{B} \right) .
		\end{aligned}
	\end{equation}
	
	Suppose we have synchronized inertial measurements in time $ t \in \{1,\cdots,T\} $. Variables to be estimated are $ {}^{B} \mathbf{R}_{A} $ and $ {}^{A} \mathbf{p}_{B} $ to complete the extrinsic calibration process.
	
	\subsection{Calibration Solution Approach}
	To prevent the Gimbal lock problem with Euler angle and computation complexity with rotation matrix, we choose the quaternion $ {}^{B} \mathbf{q}_{A} $ to represent relative orientation.
	
	We define the system state:
	\begin{equation}
		\begin{aligned}
			\mathcal{X} = \{ {}^{B} \mathbf{q}_{A}, {}^{A} \mathbf{p}_{B}  \}.
		\end{aligned}
	\end{equation}
	
	Before we optimize, assume we have known inertial intrinsic parameters. We do not evaluate gyroscope misalignment since we incorporate it into the relative rotation.
	
	Our proposed non-linear least-squares optimization problems consist of two steps: relative orientation and then translation estimations. On the one hand, the orientation calibration problem is independent of linear accelerations. The translation calibration should not pose excessive restrictions on orientation updates. On the other hand, the translation calibration problem suffers from less noise \added{than using raw inertial measurements} because of the optimized orientation parameters and angular velocities. Besides, the overall computation time is reduced.
	
	Firstly, define the orientation-related non-linear least-squares problem:
	\begin{equation}
		\begin{aligned}
			\min_{{}^{B} \mathbf{q}_{A}} \sum_{t=1}^{T} || \mathbf{r}_{\boldsymbol{\omega}}^t(\mathcal{X} ; {}^{A}\widetilde{\boldsymbol{\omega}}(t), {}^{B}\widetilde{\boldsymbol{\omega}}(t) ||_{\Sigma_{\boldsymbol{\omega}}^t}^2
		\end{aligned}
	\end{equation}
	where $ \mathbf{r}_{\boldsymbol{\omega}}^t $ is the residual related to the angular velocity measurements,
	\begin{equation}
		\begin{aligned}
			\mathbf{r}_{\boldsymbol{\omega}}^t = {}^{B}\widetilde{\boldsymbol{\omega}}(t) - {}^{B} \mathbf{q}_{A} {}^{A}\widetilde{\boldsymbol{\omega}}(t).
		\end{aligned}
	\end{equation}
	
	The corresponding covariance matrix is
	\begin{equation}
		\Sigma_{\boldsymbol{\omega}}^t = \left( (\sigma_{gA}^2 + \sigma_{gB}^2)/\Delta t + (\sigma_{b_{gA}}^2 + \sigma_{b_{gB}}^2) \Delta t \cdot t \right) \mathbf{I}_3.
	\end{equation}
	
	Secondly, define the translation-related non-linear least-squares problem:
	\begin{equation}
		\begin{aligned}
			\min_{{}^{B} \mathbf{p}_{A}} \sum_{t=1}^{T} || \mathbf{r}_{\boldsymbol{a}}^t\left(\mathcal{X} ; {}^{A}\widetilde{\boldsymbol{\omega}}(t), {}^{B}\widetilde{\boldsymbol{\omega}}(t), {}^{A}\widetilde{\boldsymbol{a}}(t), {}^{B}\widetilde{\boldsymbol{a}}(t) \right) ||_{\Sigma_{\boldsymbol{a}}^t}^2
		\end{aligned}
	\end{equation}
	where $ \mathbf{r}_{\boldsymbol{a}}^t $ is the residual regarding the inertial measurements
	\begin{equation}
		\begin{aligned}
			\mathbf{r}_{\boldsymbol{a}}^t = {}^{B}\widetilde{\boldsymbol{a}}(t) - {}^{B} \mathbf{q}_{A} \big( & {}^{A}\widetilde{\boldsymbol{a}}(t) + \lfloor {}^{A} \widetilde{\boldsymbol{\omega}}(t) \rfloor^2 {}^{A} \mathbf{p}_{B} \\ 
			& + \lfloor {}^{A}\dot{\widetilde{\boldsymbol{\omega}}}(t) \rfloor {}^{A} \mathbf{p}_{B} \big).
		\end{aligned}
	\end{equation}
	
	Because IMU cannot measure angular accelerations $ \dot{\boldsymbol{\omega}} $, instead of updating them online, we can estimate them via the time derivative of the virtual angular velocity model
	\begin{equation}
		\begin{aligned}
			{}^{A}\dot{\widetilde{\boldsymbol{\omega}}}(t) = \frac{\texttt{freq}}{4} \big(& {}^{B} \mathbf{q}_{A}^{-1} {}^{B} \widetilde{\boldsymbol{\omega}}(t + 1) - {}^{B} \mathbf{q}_{A}^{-1} {}^{B} \widetilde{\boldsymbol{\omega}}(t - 1) \\
			&+ {}^{A} \widetilde{\boldsymbol{\omega}}(t + 1) - {}^{A} \widetilde{\boldsymbol{\omega}}(t-1) \big),
		\end{aligned}
	\end{equation}
	\deleted{in which} where \texttt{freq} is the frequency of IMU. The corresponding covariance matrix is 
	\begin{equation}
		\begin{aligned}
			\Sigma_{\boldsymbol{a}}^t = \big[ & (\frac{\sigma_{gA}^2 \sigma_{gB}^2}{(\sigma_{gA}^2 + \sigma_{gB}^2)\Delta t} + \frac{\sigma_{gB}^4\sigma_{b_{gA}}^2 + \sigma_{gA}^4\sigma_{b_{gB}}^2}{(\sigma_{gA}^2 + \sigma_{gB}^2)^2} \Delta t \cdot t)^2 \\
			& + (\sigma_{aA}^2 + \sigma_{aB}^2)/\Delta t+(\sigma_{b_{aA}}^2 + \sigma_{b_{aB}}^2) \Delta t \cdot t \big] \mathbf{I}_3.
		\end{aligned}
	\end{equation}
	
	\deleted{It's clear} \added{We will show in \ref{VIRTUA  IMU GENERAL MODEL}} that the \added{virtual} gyroscope noise covariance is smaller than the original values $ \sigma_{gA}^2 / \Delta t  $ and $ \sigma_{gB}^2 / \Delta t  $, and the bias covariance also is smaller. By isolating orientation and translation estimation processes, our method saves time and requires shorter data collection. IMU noises are not estimated since we cannot impose appropriate restrictions on noises during optimization without the \deleted{ground-truth}\added{ground-true} trajectory. So our method can avoid the over-fitting problem.

	\section{VIRTUAL IMU METHOD ON MANIFOLD}
	\label{VIRTUAL IMU METHOD REVISITED}
	In this section, we give the general form of the VIMU generation method \cite{c6} and complete the proof that our proposed method performs better in relative translation calibration than using raw \deleted{accelerometers}\added{gyroscope} measurements. In order to integrate the VIMU with optimized extrinsic parameters into the VIO system, we derive the VIMU propagation on manifold.
	
	\subsection{Virtual IMU General Model}
	\label{VIRTUA  IMU GENERAL MODEL}
	In the case of two IMUs fusion, we suggest choosing the middle position of two sensors as the VIMU frame rather than picking an arbitrary pose. Because the \added{left nullspace matrix} $ \mathbf{Z}^{\boldsymbol{\top}} $ \added{\cite{c6}} \deleted{in the following} may be singular if the VIMU is much farther away from $ B $ than from $ A $ and vice versa. Give the general form of the VIMU model from two different IMUs:
	\begin{equation}
		\label{virtual imu model}
		\begin{aligned}
			{}^{V}\widetilde{\boldsymbol{\omega}} & = \mathbf{N}^{+} \widetilde{\boldsymbol{\omega}} = {}^V \boldsymbol{\omega} + \boldsymbol{b}_{gV} + \boldsymbol{\eta}_{gV} \\
			{}^{V}\widetilde{\boldsymbol{a}} &=\mathbf{T}\left(\widetilde{\boldsymbol{a}}-\mathbf{S}\left({}^{V}\widetilde{\boldsymbol{\omega}}\right)\right)
			 = {}^V \boldsymbol{a} + \boldsymbol{b}_{aV} + \boldsymbol{\eta}_{aV} - \mathbf{T} \cdot \mathbf{S}_a
		\end{aligned}
	\end{equation}
	where\added{ $\mathbf{T} $ and $\mathbf{S}(\cdot)$ are defined in \cite{c6} }. \deleted{where $\mathbf{N}^{+} $ is the Moore-Penrose inverse of $ \mathbf{N} $} \added{$ \mathbf{N}^{+} $ is}
	\begin{equation}
		\begin{aligned}
			&\mathbf{N}^{+} = \left( \mathbf{N}^{\boldsymbol{\top}} \mathbf{N} \right)^{-1} \mathbf{N}^{\boldsymbol{\top}}, \quad
			\mathbf{N} =\begin{bmatrix}
				{}^{A} \mathbf{R}_{V} / \sigma_{gA}\\
				{}^{B} \mathbf{R}_{V} / \sigma_{gB}
			\end{bmatrix} .
		\end{aligned}
	\end{equation}
	 
	 Bias and noise \deleted{terms} of virtual \deleted{angular velocity measurements} \added{gyroscope measurements} are \deleted{derived as}
		\begin{equation}
			\begin{aligned}
				& \boldsymbol{b}_{g V} = \mathbf{N}^{+} \begin{bmatrix}
					\boldsymbol{b}_{g A} / \sigma_{gA} \\
					\boldsymbol{b}_{g B} / \sigma_{gB}
				\end{bmatrix}, \quad \boldsymbol{\eta}_{g V} = \mathbf{N}^{+} \begin{bmatrix}
					\boldsymbol{\eta}_{g A} / \sigma_{gA} \\
					\boldsymbol{\eta}_{g B} / \sigma_{gB}
				\end{bmatrix} .
			\end{aligned}
		\end{equation}
	
	Thus taking expectation on both sides of Eq. \ref{virtual imu model}, the noise covariance \deleted{matrix} $ \boldsymbol{\eta}_{g V} \sim \mathcal{N}(\mathbf{0}, \mathbf{Q}_{g V}) $ and the bias covariance \deleted{matrix} $ \boldsymbol{b}_{g V} \sim \mathcal{N}(\mathbf{0}, \mathbf{Q}_{b_gV}) $ of virtual gyroscope measurements \deleted{noise} are
	\begin{equation}
		\label{Q=sigma^2/2}
		\begin{aligned}
			\mathbf{Q}_{g V} &= \mathbf{N}^{+} \mathbf{I}_3 \mathbf{N}^{+^{\boldsymbol{\top}}} =\frac{\sigma_{g A}^2 \sigma_{g B}^2}{\sigma_{g A}^2 + \sigma_{g B}^2} \mathbf{I}_3 \\
			\mathbf{Q}_{b_gV} & = \mathbf{N}^{+} 
			\begin{bmatrix}
				\sigma_{b_{gA}}^2 / \sigma_{gA}^2 \mathbf{I}_3 & \mathbf{0} \\
				\mathbf{0} & \sigma_{b_gB}^2 / \sigma_{gB}^2 \mathbf{I}_3
			\end{bmatrix}
			\mathbf{N}^{+^{\boldsymbol{\top}}} \\
		\end{aligned}
	\end{equation}
	which complete the proof that virtual angular acceleration measurements have less error. So our proposed method should have better performance in relative translation calibration. Similar derivation can be done for other terms \cite{c6}.
	
	\subsection{Virtual IMU Propagation on Manifold}
	We propose the VIMU propagation on manifold, based on Eq. \ref{virtual imu model}. The system state of the VIMU consists of orientation, position, velocity, and biases:
	\begin{equation}
		\begin{aligned}
			\boldsymbol{x}_V =
			\begin{bmatrix}
				{}^{W}\mathbf{R}_{V} &
				{}^{W}\boldsymbol{p}_{V} &
				{}^{W}\boldsymbol{v}_{V} &
				\boldsymbol{b}_{gV} &
				\boldsymbol{b}_{aV}
			\end{bmatrix}
		\end{aligned}
	\end{equation}
	
	where pose $ \begin{bmatrix} {}^{W}\mathbf{R}_{V} & {}^{W}\boldsymbol{p}_{V} \end{bmatrix} $ belongs to $ SE(3) $, and $ {}^{W}\boldsymbol{v}_{V} \in \mathbb{R}^3$ is the velocity  in the world frame. Firstly integrate the estimated bias terms into VIMU measurements:
	\begin{equation}
		\begin{aligned}
			{}^{V}\widehat{\boldsymbol{\omega}} &= {}^{V}\widetilde{\boldsymbol{\omega}} - \boldsymbol{b}_{g V} \\
			{}^{V}\widehat{\boldsymbol{a}} &= {}^{V}\widetilde{\boldsymbol{a}} - \boldsymbol{b}_{a V} + \mathbf{T} \cdot \mathbf{S}_a.
		\end{aligned}
	\end{equation}
	
	Assume the VIMU synchronizes with the camera and provides measurements at discrete time $ t $. \deleted{Thus} Between \deleted{two consecutive} keyframes at time $ i $ and $ j $, we derive the VIMU preintegration model:
	\begin{equation}
		\begin{aligned}
			\Delta \mathbf{R}_{ij} &= \prod_{t=i}^{j-1} \mathtt{Exp}\left[ \left( \widetilde{\boldsymbol{\omega}}(t) - \boldsymbol{b}_{gV}(t) - \boldsymbol{\eta}_{gV}(t) \right)\Delta t\right] \\ 
			\Delta \boldsymbol{v}_{ij} &= \sum_{t=i}^{j-1}\Delta \mathbf{R}_{it}\left(\widetilde{\boldsymbol{a}}(t) - \boldsymbol{b}_{aV}(t) - \boldsymbol{\eta}_{aV}(t) + \mathbf{T} \cdot \mathbf{S}_a \right) \Delta t  \\
			\Delta \boldsymbol{p}_{ij} &= \sum_{t=i}^{j-1} \big[ \Delta \boldsymbol{v}_{it}\Delta t + \frac{1}{2}\Delta\mathbf{R}_{it}\big(\widetilde{\boldsymbol{a}}(t) - \boldsymbol{b}_{aV}(t) \\
			&\quad \quad \quad - \boldsymbol{\eta}_{aV}(t) + \mathbf{T} \cdot \mathbf{S}_a\big) \Delta t^2 \big].
		\end{aligned}
	\end{equation}
	
	We derive the propagation equation to update the state estimate, based on the preintegration terms noise vector $\boldsymbol{\eta}_{ij}^{\Delta} = \begin{bmatrix} \delta \boldsymbol{\phi}_{ij}^{\top} & \delta \boldsymbol{v}_{ij}^{\top} & \delta \boldsymbol{p}_{ij}^{\top} \end{bmatrix}^{\top} $ and the VIMU noise vector $\boldsymbol{\eta}_t^d = \begin{bmatrix} \boldsymbol{\eta}_{gV}(t) &  \boldsymbol{\eta}_{aV}(t) \end{bmatrix}$ :
	\begin{equation}
		\label{propagation form}
		\begin{aligned}
			\boldsymbol{\eta}_{ij}^{\Delta} = \mathbf{A}_{ij-1}\boldsymbol{\eta}_{ij-1}^{\Delta} + \mathbf{B}_{j-1} \boldsymbol{\eta}_{j-1}^{d}.
		\end{aligned}
	\end{equation}
	
	From the linearized equation Eq. \ref{propagation form}, we can derive the corresponding covariance matrix:
	\begin{equation}
		\begin{aligned}
			\mathbf{\Sigma}_{ij} = \mathbf{A}_{j-1}\mathbf{\Sigma}_{ij-1}\mathbf{A}_{j-1}^{\top}  + \mathbf{B}_{j-1} \mathbf{\Sigma}_{\boldsymbol{\eta}} \mathbf{B}_{j-1}^{\top} 
		\end{aligned}
	\end{equation}
	
	where $ \mathbf{\Sigma}_{\boldsymbol{\eta}} $ is the covariance matrix of virtual IMU noise. $ \mathbf{\Sigma}_{ij} $ starts from initial value $ \mathbf{\Sigma}_{ii} = \mathbf{0}_{9 \times 9} $, and $ \mathbf{\Sigma}_{\boldsymbol{\eta}} $ starts from $ \begin{bmatrix} \mathbf{Q}_{g V} & \mathbf{0} \\ \mathbf{0} & \mathbf{Q}_{a V} \end{bmatrix} $. The detailed form of Eq. \ref{propagation form} is in Eq. \ref{propagation details}
	\begin{equation}
		\label{propagation details}
		\begin{aligned}
			\mathbf{A}_{j-1} &= \begin{bmatrix}
				\Delta \widetilde{\mathbf{R}}_{j-1j}^{\top} & \mathbf{0} & \mathbf{0} \\
				- \widetilde{\mathbf{R}}_{ij-1} \lfloor{}^{V}\widehat{\boldsymbol{a}}_{j-1}\rfloor \Delta t & \mathbf{I} & \mathbf{0} \\
				- \frac{1}{2}\widetilde{\mathbf{R}}_{ij-1} \lfloor{}^{V}\widehat{\boldsymbol{a}}_{j-1}\rfloor \Delta t^2 & \Delta t \mathbf{I} & \mathbf{I}
			\end{bmatrix} \\
			\mathbf{B}_{j-1} &= \begin{bmatrix}
				\mathbf{J}^{j-1}_r \Delta t & \mathbf{0} \\
				\widetilde{\mathbf{R}}_{ij-1} \lfloor\delta \boldsymbol{\phi}_{ij-1}\rfloor \mathbf{T} \boldsymbol{\Psi} \Delta t & \Delta \widetilde{\mathbf{R}}_{ij-1} \Delta t \\
				- \frac{1}{2} \widetilde{\mathbf{R}}_{ij-1} (\mathbf{I} - \lfloor\delta \boldsymbol{\phi}_{ij-1}\rfloor) \mathbf{T} \boldsymbol{\Psi} \Delta t^2 & \frac{1}{2}\Delta \tilde{\mathbf{R}}_{ik} \Delta t^2 
			\end{bmatrix}
		\end{aligned}
	\end{equation}
	where we have
	\begin{equation}
		\begin{aligned}
			\boldsymbol{\Psi}=\left[\begin{array}{c}
				{ }^{A} \mathbf{R}_{V}\left(-\left\lfloor^{V} \widehat{\boldsymbol{\omega}}\right\rfloor\left\lfloor^{V} \boldsymbol{p}_{A}\right\rfloor-\left\lfloor\left\lfloor^{V} \widehat{\boldsymbol{\omega}}\right\rfloor^{V} \boldsymbol{p}_{A}\right\rfloor\right) \\
				{ }^{B} \mathbf{R}_{V}\left(-\left\lfloor^{V} \widehat{\boldsymbol{\omega}}\right\rfloor\left\lfloor^{V} \boldsymbol{p}_{B}\right\rfloor-\left\lfloor\left\lfloor^{V} \widehat{\boldsymbol{\omega}}\right\rfloor^{V} \boldsymbol{p}_{B}\right\rfloor\right)
			\end{array}\right]
		\end{aligned}
	\end{equation}and the right Jacobian matrix
	\begin{equation}
		\begin{aligned}
			\mathbf{J}_r^{k} = \mathbf{J}_r^{k}\left( ({}^{V}\boldsymbol{\omega}_k - \boldsymbol{b}_{gV}^k )\Delta t \right).
		\end{aligned}
	\end{equation}
	
	Note that although we introduce some new terms in virtual inertial measurement generation and propagation on manifold, the increased computation time is relatively short. Some terms can be finished offline, including the matrix $ \mathbf{N}^{+} $ and $ \mathbf{T} $, while the computational complexity of other terms (e.g., $ \mathbf{S}_a $, $ \boldsymbol{\Psi} $ ) is $ \mathcal{O}(N) $, where $ N $ is the number of IMUs.
	
	\section{EXPERIMENTS}
	\label{experiments}
	
	Our experiments include three parts. Firstly we test our calibration method on three different sensor boards. Then we only use inertial measurements to predict motion in the simulation environment. Lastly, we integrate our proposed method into the VIO system to compare localization accuracy and system robustness.
	
	\subsection{Real-World Extrinsic Calibration Experiments}
	\begin{table*}[b]
		\setlength{\tabcolsep}{3pt}
		\caption{Calibration RMSE in Position and Orientation, Computation Time regarding Kalibr \cite{c16}, Method A \cite{c19} and Our Method.}
		\label{extrinsic_calibration_results}
		\centering
		\begin{tabular}{@{\extracolsep{4pt}}llcccccccccccc@{}}
			\noalign{\hrule height 0.5pt}
			&                 & \multicolumn{6}{c}{Datasets \cite{c19}}                                                                              & \multicolumn{2}{c}{Our Sensor Board}    & \multicolumn{2}{c}{T265 \& D435i Board}  & \multicolumn{2}{c}{Simulation}    \\ \clineB{3-8}{0.25} \clineB{9-10}{0.25} \clineB{11-12}{0.25} \clineB{13-14}{0.25}
			&                 & \multicolumn{2}{c}{Ill-Lit}        & \multicolumn{2}{c}{Blurry}         & \multicolumn{2}{c}{Baseline}       & \multicolumn{2}{c}{}                    & \multicolumn{2}{c}{}        & \multicolumn{2}{c}{}            \\
			&                 & 60s                & 2s            & 60s                & 2s            & 60s                & 2s            & 60s                & 2s                 & 60s                & 2s      & 60s                & 2s           \\ \noalign{\hrule height 0.25pt}
			\multirow{3}{*}{Kalibr}                                                                       & pos. err. (mm)  & 107.42             & N             & 70.41              & N             & 4.39               & N             & 5.07               & X                  & 37.03              & X       & N & N           \\
			& rot. err. (deg) & 1.81               & N             & 26.98              & N             & 1.35               & N             & 0.88               & X                  & 38.7               & X      & N  & N            \\
			& time (ms)            & 7.83               & N             & 98.99              & N             & 106.96             & N             & \textgreater{}1000 & X                  & \textgreater{}1000 & X    & N & N              \\ \noalign{\hrule height 0.25pt}
			\multirow{3}{*}{\begin{tabular}[c]{@{}l@{}}Method A:\\ Online Bias\\ Estimation\end{tabular}} & pos. err. (mm)  & 2.38               & 3.33          & 3.3                & 3.69          & 2.72               & 2.76          & 107.73             & 72.46              & 48.83              & 30.2    &  23.95  & 247.36          \\
			& rot. err. (deg) & 0.68               & 0.66          & 0.67               & 0.88          & \textbf{0.61}      & 0.83          & 19.56              & 1.69               & 0.72               & 0.91    & 0.04    & 0.11       \\
			& time (ms)           & \textgreater{}1000 & 57.62         & \textgreater{}1000 & 52.44         & \textgreater{}1000 & 54.49         & \textgreater{}1000 & \textgreater{}1000 & \textgreater{}1000 & \textgreater{}1000 & \textgreater{}1000 & 639.15 \\ \noalign{\hrule height 0.25pt}
			\multirow{3}{*}{\begin{tabular}[c]{@{}l@{}}Method A:\\ Const Bias\\ Hypothesis\end{tabular}}  & pos. err. (mm)  & 3.61               & 3.45          & 2.87               & 2.6           & 2.95               & 2.57          & 51.61              & 12.45              & 42.26              & 35.83      & 18.94  & 18.08        \\
			& rot. err. (deg) & 0.68               & 0.66          & 0.66               & 0.88          & 0.61               & 0.78          & 3.56               & 2.02               & 0.7                & 3.38     & 0.02    & 0.06      \\
			& time (ms)           & 78.38              & 2.86          & 67.68              & 2.63          & 78.14              & 2.42          & 338.88             & 8.64               & 99.24              & 19.99    & 879.58 & 84.62          \\ \noalign{\hrule height 0.25pt}
			\multirow{3}{*}{Our Method}                                                                   & pos. err. (mm)  & \textbf{0.12}      & \textbf{0.73} & \textbf{1.3}       & \textbf{0.31} & \textbf{0.15}      & \textbf{0.33} & \textbf{1.33}      & \textbf{0.23}      & \textbf{0.2}       & \textbf{0.38}   & \textbf{0.09}  & \textbf{0.19}    \\
			& rot. err. (deg) & \textbf{0.43}      & \textbf{0.34} & \textbf{0.55}      & \textbf{0.14} & 0.64               & \textbf{0.64} & \textbf{0.15}      & \textbf{1.15}      & \textbf{0.7}       & \textbf{0.72} & \textbf{0.01}   & \textbf{0.01}  \\
			& time (ms)           & \textbf{24.2}      & \textbf{1.31} & \textbf{24.5}      & \textbf{1.17} & \textbf{24.78}     & \textbf{1.49} & \textbf{84.08}     & \textbf{2.35}      & \textbf{52.54}     & \textbf{3.42}   & \textbf{62.2} & \textbf{16.96}   \\ \noalign{\hrule height 0.5pt}
		\end{tabular}
		\begin{tablenotes}
			\footnotesize
			\item[] N means results are unavailable because of unknown camera intrinsic parameters.
			\item[] X means Kalibr optimization failure.
		\end{tablenotes}
	\end{table*}

\begin{figure}[thpb]
		\centering
		\includegraphics[width=0.3\textwidth]{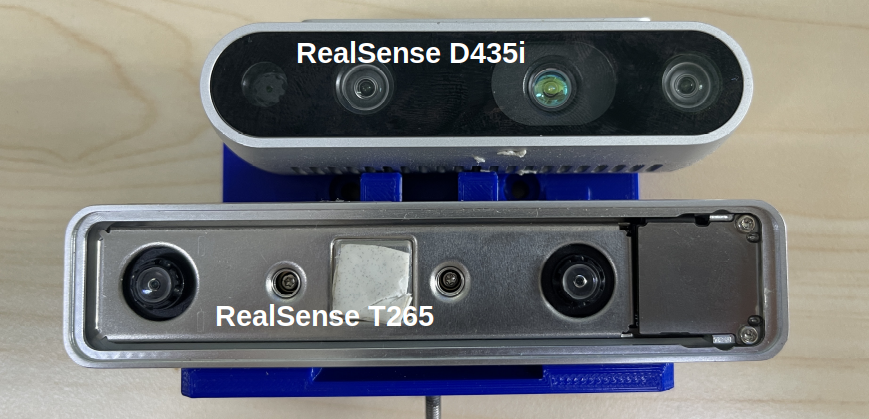}
		\caption{The RealSense T265 and D435i sensor board.}
		\label{t265d435i}
	\end{figure}

	In this section, we test extrinsic calibration on the dataset in \cite{c19}, our self-made sensor board (see Fig. \ref{leadsense}), and a mount with RealSense T265 and D435i sensors (see Fig. \ref{t265d435i}), compared with Kalibr and the method in \cite{c19}. We use "method A" as the method in \cite{c19}.
	
	The datasets in \cite{c19} have three environments, including "baseline", "ill-lit" and "blurry". Our self-made sensor board includes a pair of stereo cameras, providing synchronized gray images of $ 1280 \times 720 $ at 50 Hz. It also has a pair of  MPU 6050 inertial sensors at 250 Hz. Since each IMU is placed below the corresponding camera based on the PCB design, we assume the relative position is identical to the camera baseline of 120.32 mm, and the relative orientation angle is near 0 degrees. In the third experiment, RealSense T265 outputs $848 \times 800$ stereo images at 30 Hz and inertial measurements at 200 Hz. RealSense D435i outputs inertial measurements at 400 Hz. The ground-true extrinsic parameters are derived from factory-calibrated camera-IMU parameters and the 3D printed mount size. In each sequence, we use the entire dataset and extract two seconds of data to compare calibration performances.
	
	Results are shown in Table \ref{extrinsic_calibration_results}, including the root-mean-square error (RMSE) in relative position, orientation angle, and computation time. We use a laptop with 32 GB RAM and Intel i7-11800H CPU operating at 2.30 GHz, and we allocate eight threads. Firstly, in the datasets \cite{c19} experiment, we keep the results of Kalibr since we do not know the camera intrinsic parameters. We take the mean result value in each environment. Results validate that our method is better than Kalibr and method A if we use the entire measurements. Furthermore, the benefit of our method is more apparent when extracting a few data (e.g., two seconds long).
	
	Secondly, method A fails with our self-made and the T265 with D435i sensor boards. Meanwhile, our method succeeds with some sharp movements in each trajectory. In conclusion, our method outperforms other competing methods in the accuracy, computation time, and robustness.

	\subsection{Simulation Experiments of IMU Integration}
\begin{figure}[htbp]
		\centering
		\includegraphics[width=0.22\textwidth]{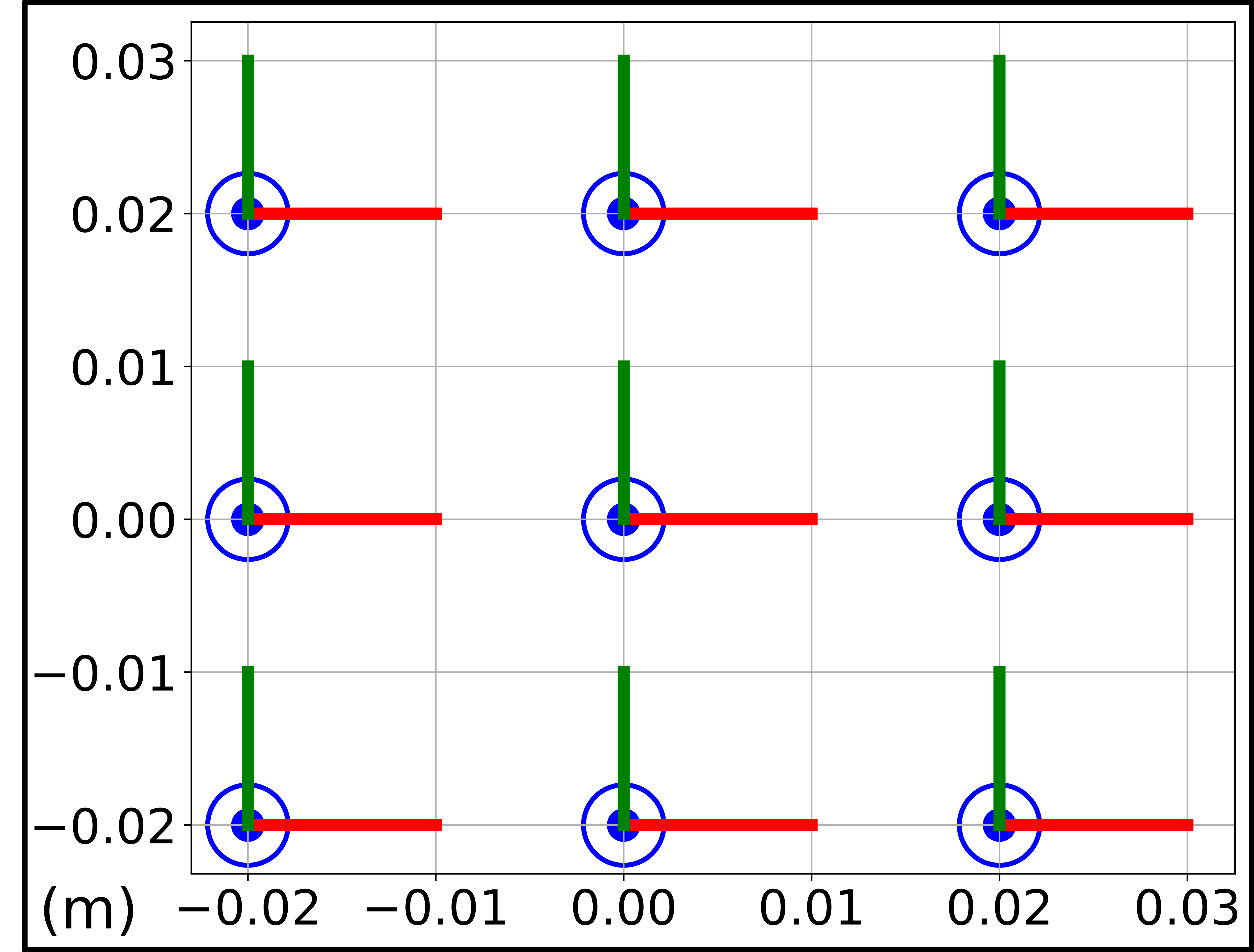}
		\caption{The simulated IMU array. [Red]: X axis; [Green]: Y axis; [Blue]: Z axis.}
		\label{simulation imu array}
	\end{figure}

	We use an open-source tool\footnote{ \href{https://github.com/HeYijia/vio_data_simulation}{https://github.com/HeYijia/vio\_data\_simulation} } to simulate the nine-IMU array measurements in arbitrarily user-defined trajectory. Fig. \ref{simulation imu array} shows the coordinate of each IMU.
	
	\deleted{Assume} \added{The simulated IMU array poses are added} \deleted{the extrinsic parameters have} \added{with} rotation and translation errors that all obey Gaussian distribution with zero mean\deleted{error}. \deleted{Specifically,}The standard deviation is 0.01 radians for rotation error and 0.001 meters for translation error. \added{We use the same IMU bias and noise parameters as our sensor board.}

	During data generation, inertial measurements are generated at 200 Hz and synchronized. Then we sample 100 different extrinsic errors based on the Gaussian distribution; for each, we generate 5000 sequences to compare the root-mean-square error (RMSE) in position, orientation angle, and velocity. \added{Table \ref{extrinsic_calibration_results} shows the extrinsic calibration results based on the simulated ground-true parameters. Our method performs better in the accuracy and computation time.}
	
\begin{figure}[t]
		\vspace*{0.08in}
		\centering
		\includegraphics[width=0.38\textwidth]{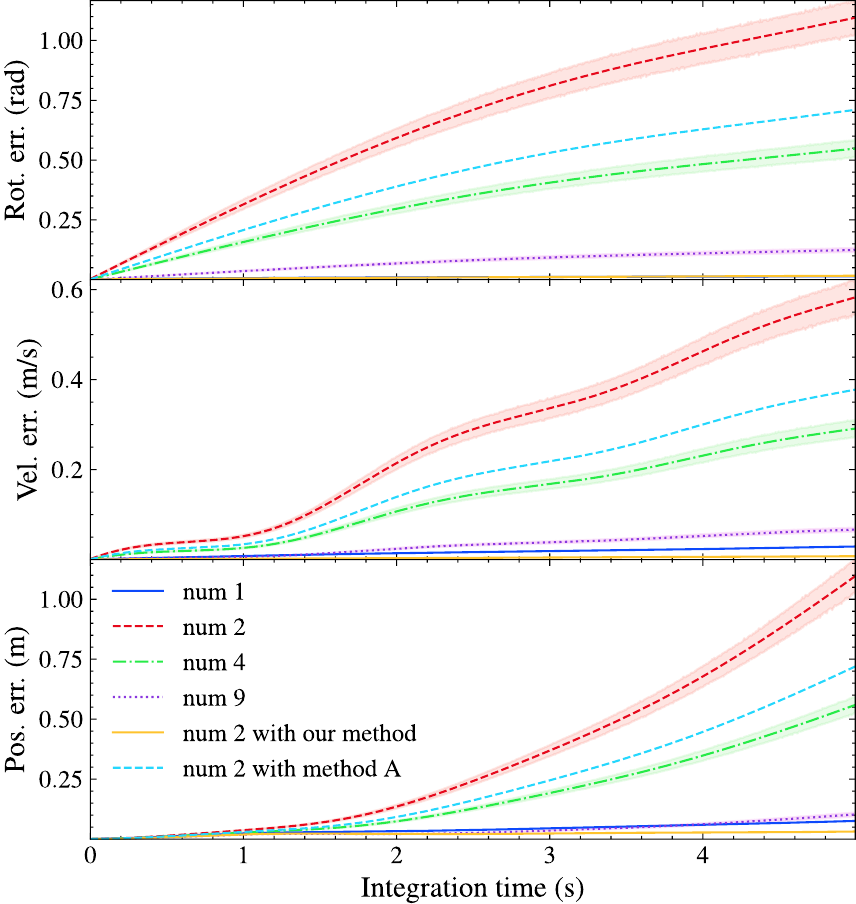}
		\caption{RMSE in velocity, orientation angle, and position with IMU preintegration. The numbers 2, 4, and 9 have tiny extrinsic errors, plotted with the error bar. [Top]: predicted velocity RMSE; [Middle]: predicted orientation angle RMSE; [Bottom]: predicted position RMSE.}
		\label{RMSE error of pvq}
	\end{figure}

	The result in Fig. \ref{RMSE error of pvq} shows different performances of two, four, and nine IMUs with tiny extrinsic errors, one IMU with ground-true extrinsic parameters, and two IMUs with our calibration method and method A\deleted{, respectively}. With the number of IMUs fused increasing, the \added{VIMU} integration result \deleted{of virtual inertial measurements} becomes more precise. \added{But}\deleted{However,} results show that \deleted{using}one IMU rather than MIMU with tiny extrinsic errors provides better motion estimation. Our proposed calibration method is proven to improve the inertial integration accuracy. Even two IMUs with optimized extrinsic parameters can perform better than nine.
	
	\subsection{VIO Experiments on OpenLORIS Benchmark}
	In this section, we integrate our proposed method into the visual-inertial system ORB-SLAM3 \cite{c8} with no loop closure, experimenting on the OpenLORIS-Scene dataset \cite{c26}.
	We use the two fisheye cameras of RealSense T265 as imaging sensors in all experiments. For the two-IMU setup, we fuse two IMUs of RealSense T265 and D435i, using the extrinsic parameters estimated from Kalibr in \cite{c26}. We choose the IMU of RealSense T265 for the one-IMU setup. We extract 2 seconds of data to do \deleted{the} extrinsic calibration \deleted{in the setup of two-IMU} with our method.
	
	Table \ref{openloris_results} shows that the stereo two-IMU system may not outperform the stereo one-IMU system in all sequences. Even if tiny extrinsic errors exist, the inertial preintegration accuracy can be significantly reduced. Our method can enhance the VIMU fusion performance by directly estimating precise extrinsic parameters between inertial sensors, thus increasing the visual-inertial localization accuracy.
	
	\begin{table}[t]
		\vspace*{0.08in}
		\setlength{\tabcolsep}{5pt}
		\caption{OpenLORIS-Scene Benchmark: RMSE ATE (cm) for Scenes with Ground-True Trajectory.}
		\label{openloris_results}
		\centering
		\begin{tabular}{lcccc}
			\noalign{\hrule height 0.5pt}
			\multicolumn{1}{c}{} & Stereo    & \begin{tabular}[c]{@{}c@{}}Stereo\\ 1-IMU\end{tabular} & \begin{tabular}[c]{@{}c@{}}Stereo\\ 2-IMU\end{tabular} & \begin{tabular}[c]{@{}c@{}}Stereo 2-IMU\\ with Our Method\end{tabular} \\ \noalign{\hrule height 0.25pt}
			office1-1                  & 5.9037    & 5.8042                                                 & 5.3991                                                  & \textbf{4.9954}                                                         \\
			office1-2                  & 7.4245    & 6.8589                                                 & 6.7129                                                 & \textbf{6.6698}                                                                  \\
			office1-3                  & 4.7924    & 0.5313                                                 & 0.504                                                   & \textbf{0.4783}                                                         \\
			office1-4                  & 43.2143   & 9.0734                                                 & 8.9372                                                  & \textbf{8.0642}                                                         \\
			office1-5                  & 23.2672   & 23.1459                                                & 23.1078                                                 & \textbf{22.9416}                                                        \\
			office1-6                  & 38.0016   & 6.524                                                  & 6.4621                                                  & \textbf{6.2492}                                                         \\
			office1-7                  & 31.1784   & 5.8838                                                 & 5.8247                                                  & \textbf{5.67}                                                         \\ 
			market1-1                  & X         & 188.5224                                               & 120.1273                                                & \textbf{91.7784}                                                        \\
			market1-2                  & X         & 126.0484                                               & 117.473                                                 & \textbf{110.482}                                                        \\
			market1-3                  & X         & 140.7409                                               & 142.3443$\downharpoonright$                                                & \textbf{135.1464}                                                       \\ 
			corridor1-1                & X         & 93.4954                                                & 117.7187$\downharpoonright$                                                & \textbf{77.9467}                                                        \\
			corridor1-2                & X         & 47.2323                                                & 50.437$\downharpoonright$                                                  & \textbf{45.3384}                                                        \\
			corridor1-3                & 1134.4583 & 771.6293                                               & 764.6611                                                & \textbf{758.9993}                                                       \\
			corridor1-4                & X         & 15.7554                                                & 16.755$\downharpoonright$                                                  & \textbf{14.1547}                                                        \\
			corridor1-5                & X         & 64.0874                                                & 66.4263$\downharpoonright$                                                 & \textbf{55.1908}                                                        \\ 
			home1-1                    & X         & 39.0209                                                & 38.3991                                                 & \textbf{36.3828}                                                        \\
			home1-2                    & X         & 32.8175                                                & 32.7885                                                 & \textbf{32.0464}                                                        \\
			home1-3                    & X         & 34.6824                                                & 33.3908                                                 & \textbf{33.3801}                                                        \\
			home1-4                    & X         & 33.9653                                                & 32.1762                                                 & \textbf{30.6295}                                                        \\
			home1-5                    & 66.5787   & 26.5857                                                & 26.8888$\downharpoonright$                                          & \textbf{17.9694}                                                         \\
			cafe1-1                    & X         & 10.6222                                                & 10.7999$\downharpoonright$                                                 & \textbf{10.0917}                                                        \\
			cafe1-2                    & X         & 11.3821                                                & 11.0009                                                 & \textbf{10.6747}                                                        \\ \noalign{\hrule height 0.5pt}
		\end{tabular}
		\begin{tablenotes}
			\footnotesize
			\item[] X means stereo tracking failure.
			\item[] $\downharpoonright$ means the localization accuracy of the stereo two-IMU set is worse than that of the stereo one-IMU set in the same sequence.
		\end{tablenotes}
	\end{table}
	
	\subsection{Real-World VIO Experiments}
	For the completeness of VIO experiments, we integrate our method into ORB-SLAM3 with no loop closure, using datasets collected with our self-made sensor board.
	
\begin{figure}[htbp]
		\vspace*{0.08in}
		\centering
		\includegraphics[width=0.38\textwidth]{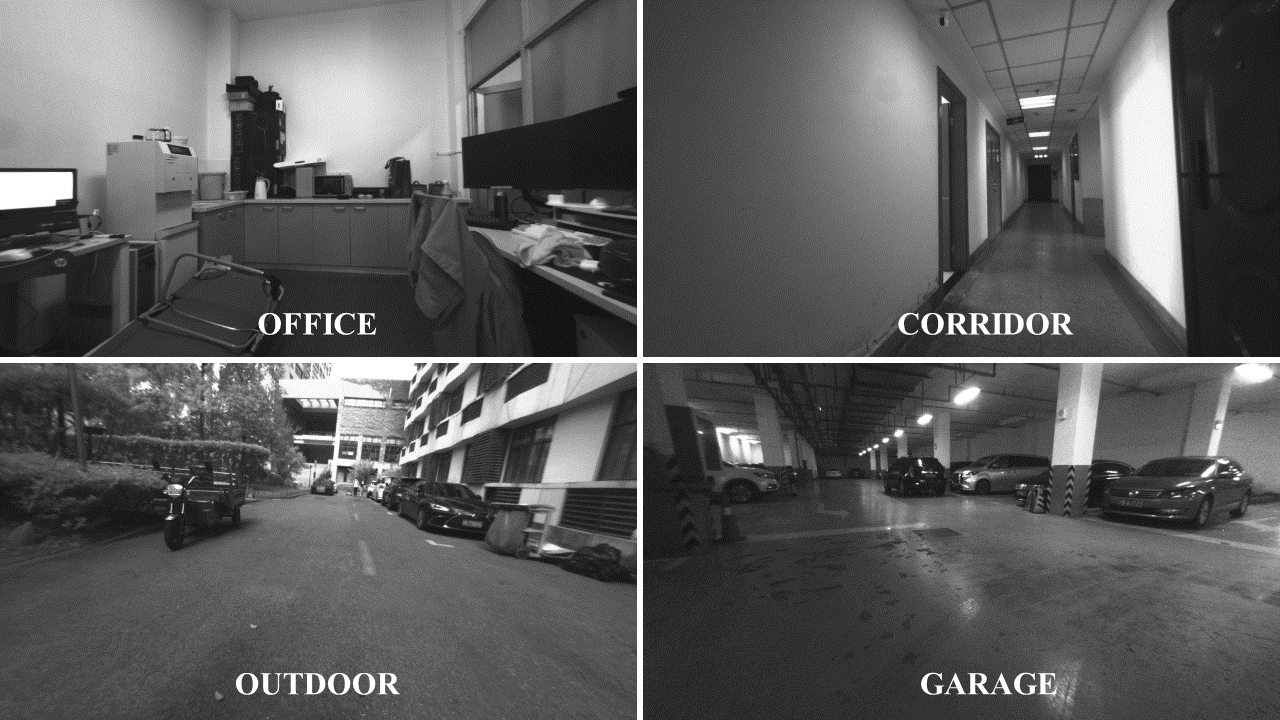}
		\caption{Examples of images in real-world experiments  captured with our self-made sensor board.}
		\label{real-world example figs}
	\end{figure}

	
	First we use Leica Nova TS60 \footnote{\href{https://leica-geosystems.com/products/total-stations/robotic-total-stations/leica-nova-ts60}{https://leica-geosystems.com} } with ground-true position to record datasets in the \texttt{office}, \texttt{corridor}, \texttt{outdoor}, and \texttt{garage}. Example pictures are shown in Fig. \ref{real-world example figs}. We randomly add short-term sharp movements in all sequences.
	Then use the OptiTrack \footnote{\href{https://www.optitrack.com}{https://www.optitrack.com} } to record datasets with ground-true position and orientation. Details are as follows:
	
	1) The trajectories recorded with the vehicle are smooth, while handheld datasets have continuous shakes and randomly short-term sharp movements.
	
	2) In \texttt{handheld 1} cameras move along a regular circular route; in \texttt{handheld 2} cameras move along the same way but observe from opposite views; in \texttt{handheld 3} the environment is an open area; in \texttt{handheld 4} we have dynamic objects including people and other vehicles; in \texttt{handheld 5} cameras suffer from significant illumination changes. The same pattern holds for the \texttt{vehicle 1-5}.
	
	\begin{table}[t]
		\vspace*{0.08in}
		\setlength{\tabcolsep}{5pt}
		\caption{RMSE ATE (cm) for Our Sensor Board with Ground-True Trajectory.}
		\label{leadsense real-world result}
		\centering
			\begin{tabular}{lcccc}
				\hline
				\multicolumn{1}{c}{Seq.} & Stereo   & \begin{tabular}[c]{@{}c@{}}Stereo\\ 1-IMU\end{tabular} & \begin{tabular}[c]{@{}c@{}}Stereo\\ 2-IMU\end{tabular} & \begin{tabular}[c]{@{}c@{}}Stereo 2-IMU\\ with Our Method\end{tabular} \\ \hline
				corridor                   & 458.9071 & 276.7324                                               & 275.9376                                               & \textbf{266.8863}                                                      \\
				office                     & 99.8401  & 71.886                                                 & 71.6695                                                & \textbf{70.1605}                                                       \\
				outdoor                    & X        & 350.3261                                               & 349.43                                                 & \textbf{341.249}                                                       \\
				garage                     & 441.399  & 403.0967                                               & 400.5042                                               & \textbf{379.1043}                                                      \\ 
				handheld 1                 & 98.8325  & 97.0279                                                & 77.2796                                                & \textbf{71.1463}                                                       \\
				handheld 2                 & 51.2822  & 38.0673                                                & 38.2407                                                & \textbf{34.3224}                                                       \\
				handheld 3                 & 89.0383  & 37.3369                                                & 36.9495                                                & \textbf{32.8598}                                                       \\
				handheld 4                 & 109.4845 & 41.8153                                                & 41.8329                                                & \textbf{38.5077}                                                       \\
				handheld 5                 & 57.1734  & 29.6313                                                & 32.1108                                                & \textbf{26.1939}                                                       \\ 
				vehicle 1                  & 50.0702  & XX                                                      & 18.2564                                                & \textbf{16.5809}                                                       \\
				vehicle 2                  & 85.1862  & 38.5989                                                & 39.7027                                                & \textbf{35.6422}                                                       \\
				vehicle 3                  & 162.9419 & 31.1185                                                & 29.4017                                                & \textbf{26.6551}                                                       \\
				vehicle 4                  & 56.6178  & 51.9887                                                & 48.1439                                                & \textbf{43.3746}                                                       \\
				vehicle 5                  & 39.7138  & 29.3192                                                & 25.8514                                                & \textbf{23.7524}                                                       \\ \hline
			\end{tabular}
		\begin{tablenotes}
			\footnotesize
			\item[] X means stereo tracking failure.
			\item[] XX means IMU initialization failure.
		\end{tablenotes}
	\end{table}
	
	Table \ref{leadsense real-world result} shows that our proposed calibration method can improve VIO localization accuracy through extrinsic precision lifting. We do not compare with Kalibr and method A since they both output false extrinsic parameters. In addition to the VIMU method, our proposed method can be integrated into other IMU fusion algorithms, VIO, and LiDAR inertial odometry (LIO) systems.
	
	

	\section{CONCLUSIONS}
	In this paper, we propose a fast extrinsic calibration method between MIMU. We first estimate the relative orientation and then improve the relative translation accuracy by introducing the VIMU method. The proposed method is validated to be fast, precise, robust, and free from ground-true trajectory, external sensors, and noises online estimation. Then we give the general form of the VIMU method and propose the VIMU propagation on manifold. The experiment results show that our proposed calibration method can improve VIO localization accuracy.

	
	

	

\end{document}